\documentclass[letterpaper]{article} 
\usepackage{aaai2027}  
\usepackage[hyphens]{url}  
\usepackage{graphicx} 
\usepackage{natbib}  
\usepackage{caption} 
\usepackage{algorithm}
\usepackage{algorithmic}
\usepackage{amsmath}
\usepackage{amssymb}
\usepackage{bm}
\usepackage{newfloat}
\usepackage{listings}
\DeclareCaptionStyle{ruled}{labelfont=normalfont,labelsep=colon,strut=off} 
\floatstyle{ruled}
\newfloat{listing}{tb}{lst}{}
\floatname{listing}{Listing}

\usepackage{booktabs}

\usepackage{bibentry}
\usepackage{booktabs}   
\usepackage{multirow}   
\usepackage[table]{xcolor} 

\title{FARCA: Fact-Aligned Reliability-Aware Credit Assignment \\for Reinforcement Learning with Factual Supervision}

\nocopyright
\author {
    Qiming Xie\textsuperscript{\rm 1},
    Wenjie Zheng\textsuperscript{\rm 1},
    Xiangqing Shen\textsuperscript{\rm 2},
    Rui Xia\textsuperscript{\rm 2}\corresponding
}
\affiliations {
    \textsuperscript{\rm 1}School of Computer Science and Engineering,\\
Nanjing University of Science and Technology, Nanjing, China\\
    \textsuperscript{\rm 2}School of Intelligence Science and Technology, Nanjing University, China\\
    \texttt{\{qmxie, wjzheng\}@njust.edu.cn} \quad \texttt{\{xqshen, rxia\}@nju.edu.cn}
}

\begin{document}

\maketitle

\begin{abstract}
To reduce the hallucination risk caused by outcome-driven rewards in large language models trained through reinforcement learning with verifiable rewards, existing mitigation approaches introduce process-level factual supervision. However, due to coarse-grained aggregation of factual signals and the lack of reliability assessment for these signals, they create a mismatch between fact verification and policy updates. We term this noisy factual credit assignment and decompose it into two aspects: credit localization ambiguity and credit reliability ambiguity.
To address these issues, we propose FARCA (Fact-Aligned Reliability-Aware Credit Assignment), a policy optimization framework that transforms factual supervision into localized, reliability-weighted token-level training signals. FARCA achieves fine-grained credit localization by aligning the granularity of fact verification with that of policy updates. It further introduces counterfactual evidence attribution, which uses the dependence of a factual judgment on key evidence as an empirical proxy for verification reliability to compute reliability weights. These weights modulate factual rewards and local policy advantages, reducing the influence of potentially unreliable signals on policy optimization.
Experiments across different models and multiple factual reasoning benchmarks show that FARCA significantly improves model factuality while preserving general reasoning capabilities.
\end{abstract}

\begin{figure}[t]
    \centering
    \includegraphics[width=0.97\columnwidth]{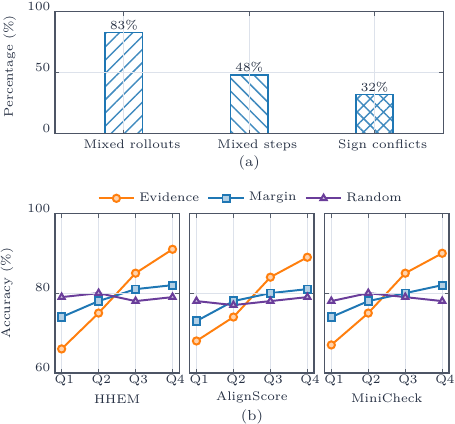}
    \caption{
    (a) Percentages of rollouts and reasoning steps with mixed-correctness atomic facts, and of sign conflicts between fact labels and their shared credit.
    (b) Accuracy of three verifiers across quartiles constructed separately for each  reliability proxy.
    }
    \label{fig:motivation}
\end{figure}

\section{Introduction}

Driven by the new paradigm of Reinforcement Learning with Verifiable Rewards (RLVR), Large Language Models (LLMs) have recently shown strong capabilities on complex tasks~\cite{jaech2024openai,guo2025deepseek}. However, prior work has shown that outcome-reward-driven RL can lead models to reach correct answers via factually flawed reasoning in knowledge-intensive tasks, failing to penalize and even reinforcing such behavior, thereby amplifying hallucination~\cite{huang2025survey,zhang2025siren}.

To mitigate this issue, recent studies have introduced factual supervision into RLVR, demonstrating its promise for improving model factuality. One line of work constructs process-aware factual rewards at the trajectory level to constrain model generation~\cite{ren-etal-2026-knowrl}, while another refines the granularity of supervision by introducing factual signals at the reasoning-step level to guide advantage updates~\cite{li2025hallucination,gui2026learning}. Underlying these approaches is a common mechanism: the outcome of fact verification is converted into a signal attributed to the specific generated content responsible for it, determining the direction and magnitude of the corresponding policy update. We term this \emph{factual credit}, and as the intermediary linking fact verification with policy optimization, its assignment quality determines whether verification outcomes can be faithfully converted into policy gradients on the corresponding content. However, existing work has not adequately addressed two critical steps in this conversion: \textbf{where} credit should be assigned, and \textbf{how reliable} the corresponding verification signal is. Mishandling either step causes the factual credit that actually enters the policy objective to deviate from ideal factual supervision in location, direction, or magnitude. We refer to this deviation as \emph{noisy factual credit assignment}, which we decompose into two interrelated aspects: credit localization ambiguity and credit reliability ambiguity.

Credit localization ambiguity arises from the granularity mismatch between fact verification and policy updates: verification operates at the level of atomic facts or reasoning steps, whereas RL updates parameters via per-token log-probabilities. When a trajectory or step contains facts of mixed factuality, collapsing them into a single scalar credit blurs the correspondence between each verification outcome and its source tokens, causing tokens tied to correct and incorrect facts to share the same credit. Our pilot experiments confirm this is not a marginal issue (Figure~\ref{fig:motivation}(a)). Using reasoning trajectories from \texttt{DeepSeek-R1-Distill-Qwen-7B}~\cite{guo2025deepseek} on 100 examples sampled from 2WikiMultiHopQA~\cite{ho-etal-2020-constructing}, we find mixed correctness is common, with credit aggregation assigning a sign opposite to the true label to a substantial portion of fact spans.

Credit reliability ambiguity stems from the inherent uncertainty of verification outcomes. Existing methods typically formulate fact verification as a Natural Language Inference (NLI) task and use the verifier's output directly as supervision, without assessing its reliability. Yet NLI-based verifiers can be misled by linguistic priors or irrelevant spans, producing misjudgements that contaminate the policy gradient~\cite{chen-etal-2025-reference,cai2025reinforcement}. Intuitively, a trustworthy verification outcome should be grounded in identifiable evidence: removing key evidence should change the score accordingly. Using the pilot experiment data, we evaluate three reliability proxies against human-annotated accuracy across three verifiers with distinct architectures (HHEM~\cite{hhem-2.1-open}, AlignScore~\cite{zha-etal-2023-alignscore}, MiniCheck~\cite{tang-etal-2024-minicheck}): the score change from removing the most relevant evidence (Evidence) or random evidence (Random), and the margin from the classification boundary (Margin). As shown in Figure~\ref{fig:motivation}(b), accuracy increases monotonically with the Evidence quantile across all verifiers, supporting evidence dependence as a signal for reliability-aware credit modulation.

To this end, we propose FARCA (Fact-Aligned Reliability-Aware Credit Assignment), a unified framework for reliable factual credit assignment. \textbf{To address credit localization ambiguity}, FARCA first decomposes each trajectory into a set of independently verifiable atomic facts, each anchored to its originating token span within the trajectory, termed its token provenance. Then, it assesses each atomic fact against external evidence using a factual verifier and derives a continuous, signed factual score accordingly, routed back to the tokens identified by its provenance. By aligning the granularity between fact verification and policy updates, FARCA enables tokens corresponding to correct facts and to hallucinated content within the same reasoning process to receive differentiated optimization signals, thereby reducing the interference caused by coarse-grained aggregation of verification outcomes.
\textbf{To address credit reliability ambiguity}, FARCA employs \emph{counterfactual evidence attribution}, using the degree to which a verification outcome depends on key evidence as an empirical proxy for its reliability. Specifically, for each atomic fact, we first identify its most relevant evidence sentences, compare the verification score before and after removing this evidence, and map the resulting score change to a continuous reliability weight. This weight is further used to modulate the factual reward and the local policy advantage, limiting the influence of unreliable factual signals on policy optimization.

\textbf{Summary of Contributions.}
(1) We identify \emph{noisy factual credit assignment} as a key bottleneck in RL with factual supervision, and show that effective factual optimization requires both precise credit localization and reliable supervision signals.
(2) We propose FARCA, a unified framework that converts verification signals into localized, reliability-aware factual credit, enabling factual supervision to guide policy optimization more faithfully and robustly.
(3) Experimental results demonstrate that FARCA significantly improves model factuality while maintaining strong reasoning performance and training stability.

\section{Related Work}

\paragraph{Outcome-Driven Reinforcement Learning and Factuality.}
Reinforcement Learning with Verifiable Rewards (RLVR) has become a central paradigm for eliciting complex reasoning from large language models (LLMs), achieving substantial gains in mathematical and general reasoning tasks by optimizing automatically verifiable outcome-level rewards~\cite{shao2024grpo,guo2025deepseek}. However, these rewards assess only the correctness of the final answer and do not verify the factual accuracy of intermediate reasoning steps. This limitation is particularly problematic in knowledge-intensive tasks, where a correct answer often depends on multiple factual claims supported by external evidence. A reasoning trajectory may contain guesses or fabricated claims and still receive a positive reward if its final answer happens to be correct. As a result, outcome-based training may reinforce factually incorrect or unsupported reasoning trajectories and increase the risk of hallucinations in reasoning LLMs~\cite{ding2025fapo,paul2024making,chen2025reasoning,wang2026joint,li2025hallucination,chen2025learning}.

\paragraph{Factual Supervision in Reinforcement Learning.}
To mitigate this, recent work has incorporated factual supervision into reinforcement learning. KnowRL decomposes the reasoning process into atomic facts and aggregates the verification results into a trajectory-level factual reward~\cite{ren-etal-2026-knowrl}. FSPO performs fact verification at each reasoning step and uses the outcomes to adjust policy advantages~\cite{li2025hallucination}. FaithRL similarly verifies reasoning faithfulness at the step level and suppresses reasoning steps unsupported by evidence~\cite{gui2026learning}. These studies demonstrate the potential of incorporating factual supervision into reinforcement learning to improve model factuality.
However, these methods share factual signals within a trajectory or reasoning step, preventing them from distinguishing correct from incorrect content within the same aggregation unit and creating a granularity mismatch between fact verification and policy optimization. Moreover, they typically use verifier outputs directly to construct factual rewards or adjust advantages, overlooking the risk that verifier misjudgment may induce incorrect policy updates~\cite{chen-etal-2025-reference,cai2025reinforcement}.
FARCA addresses these limitations through fact--token alignment and reliability modeling, providing reliable, fact-guided signals for accurate and robust policy optimization.

\begin{figure*}[t]
    \centering
    \includegraphics[width=0.95\textwidth]{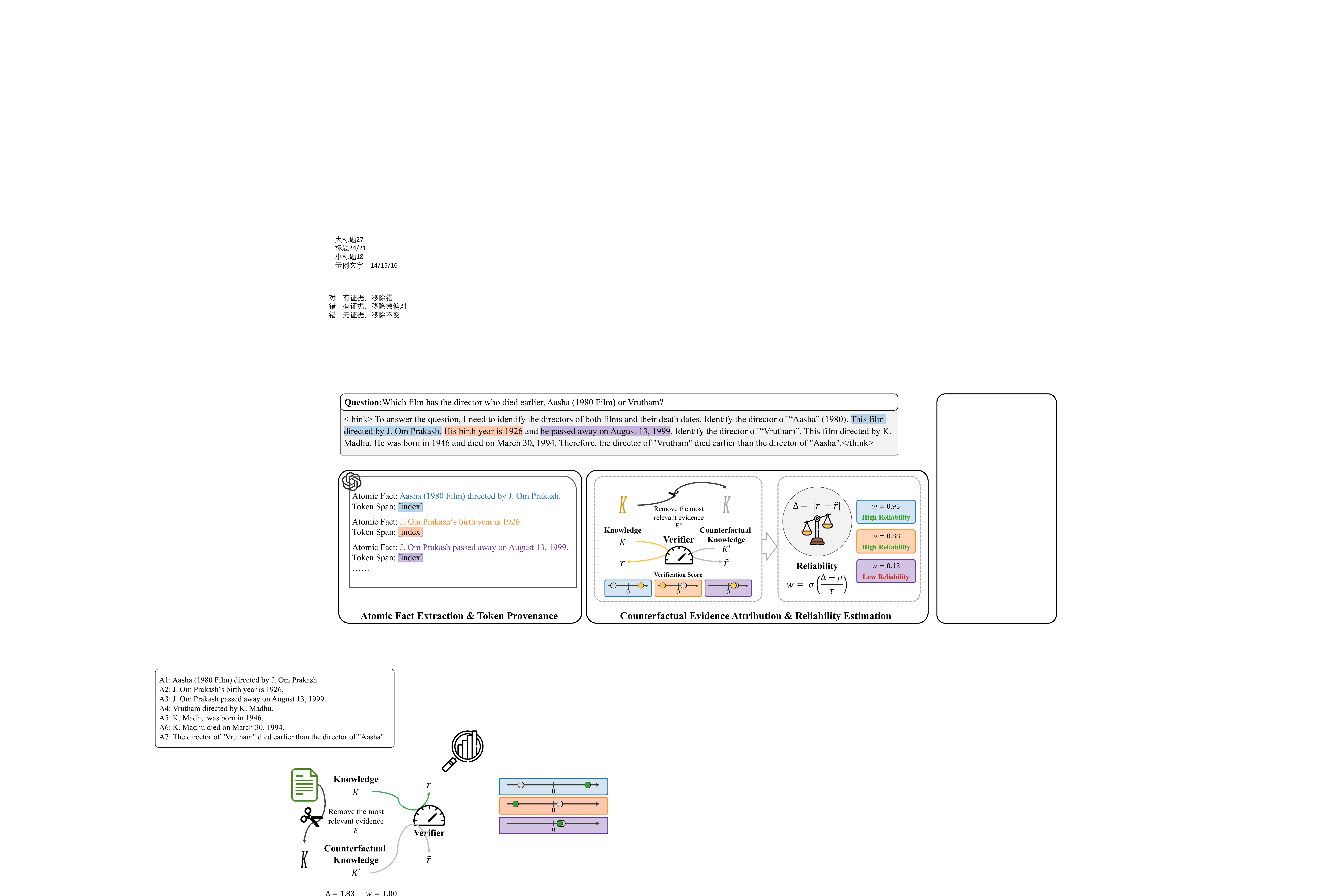}
    \caption{Overview of the proposed FARCA framework.}
    \label{fig:framework}
\end{figure*}

\section{Methodology}
\label{sec:method}
The core idea of FARCA is to achieve fact-aligned reliability-aware credit assignment.
Given a question $x$ and its corresponding knowledge snippets $K$ used as evidence, the current policy $\pi_\theta$ samples a group of rollouts $\{y_i\}_{i=1}^{G}$ for the same question. Each rollout is written as $y_i=(y_i^{\mathrm{think}},y_i^{\mathrm{answer}})$, where $y_i^{\mathrm{think}}$ denotes the intermediate reasoning text and $y_i^{\mathrm{answer}}$ denotes the answer.

\subsection{Atomic Fact Extraction and Token Provenance}
\label{sec:atomic-alignment}
FARCA first decomposes the reasoning text into atomic facts and, for each atomic fact, traces it back to its source tokens in the reasoning text. We refer to this traceable relationship as \emph{token provenance}, which determines which generation positions are ultimately supervised by the factuality signal.

Specifically, we segment the reasoning text $y_i^{\mathrm{think}}$ into sentences $y_i^{\mathrm{think}}\rightarrow\{s_{i,1},s_{i,2},\ldots,s_{i,M_i}\}$, and then filter out sentences that contain only greetings, formatting instructions, reasoning connectives, or other content without verifiable information.
For each verifiable sentence $s_{i,j}$, we prompt GPT-4o~\cite{hurst2024gpt4o} to decompose it into a set of independently verifiable atomic facts $\mathcal{C}_{i,j}=\{c_{i,j,1},c_{i,j,2},\ldots,c_{i,j,N_{i,j}}\}$, where each $c_{i,j,k}$ is a semantically self-contained claim. If the original sentence contains pronouns, ellipses, or expressions that depend on preceding context, the extraction process simultaneously performs decontextualization, rewriting it into a form that can be verified independently.
We also instruct GPT-4o to return, for each atomic fact, its source token span and index set $\mathcal{T}(c_{i,j,k})\subseteq\{1,\ldots,|y_i|\}$ in the original sentence, which we define as the token provenance of that fact.

Since the returned token provenance may be affected by repeated entities, pronoun rewriting, and coordinate structures, we apply rule-based checks and corrections to the token provenance results:
(1) For cases where the same entity appears multiple times within the same sentence, we maintain a left-to-right matching cursor to prevent an atomic fact from a later clause being incorrectly bound to an earlier occurrence of the entity.
(2) For atomic facts that underwent decontextualization, we check whether their token provenance correctly maps back to the pronoun, elliptical expression, or corresponding source span in the original sentence.
(3) For coordinate structures sharing a common prefix, we preserve the individual coverage relations of each atomic fact so that they can be handled consistently during subsequent aggregation.
(4) Atomic facts whose source span cannot be located within the original text are discarded.

This procedure yields $\mathcal{C}_i=\bigcup_{j=1}^{M_i}\mathcal{C}_{i,j}$, the set of all atomic facts extracted from the reasoning text, along with the token provenance $\mathcal{T}(c_{i,j,k})$ of each atomic fact.

\subsection{Atomic Fact Verification}
\label{sec:fact-verification}

Following prior work, we formulate fact verification as the NLI task. Given a knowledge snippet $K$ as evidence and an atomic fact $c_{i,j,k}$, the verifier outputs a score indicating how well the fact is supported by the evidence, $h_{i,j,k}=\mathrm{Verifier}(K,c_{i,j,k})\in[0,1]$. Rather than thresholding the verifier output into a discrete score, FARCA linearly maps it onto a signed interval to obtain a continuous signed factual score $r_{i,j,k}=2h_{i,j,k}-1\in[-1,1]$. $r_{i,j,k}>0$ is positive when the evidence supports the fact and negative when it contradicts the fact; a value of zero indicates a neutral judgement. Compared with a discrete factual score, the continuous signed factual score preserves the verifier's support strength, providing a smoother factual supervision signal for the subsequent reward and advantage reshaping.

\subsection{Counterfactual Evidence Attribution and Reliability Estimation}
\label{sec:counterfactual-reliability}

FARCA uses the degree to which a verification outcome depends on key evidence as an empirical proxy for its reliability, computing a reliability weight for each factual signal via counterfactual evidence attribution.

Specifically, we split the evidence into sentences $K=\{e_1,e_2,\ldots,e_L\}$, and then use a sentence encoder $\phi(\cdot)$ to compute the semantic similarity between each evidence sentence and the atomic fact, selecting the $K_{\mathrm{rel}}$ most relevant evidence sentences $\mathcal{E}^{\star}_{i,j,k}=\operatorname*{TopK}_{e_\ell\in K}^{K_{\mathrm{rel}}}\cos\!\big(\phi(e_\ell),\phi(c_{i,j,k})\big)$. These sentences are treated as the evidentiary anchors on which the current verification outcome most likely depends.

We then remove them from the original evidence to construct a counterfactual evidence set $K'_{i,j,k}=K\setminus\mathcal{E}^{\star}_{i,j,k}$, and re-verify to obtain a new score $\tilde{h}_{i,j,k}=\mathrm{Verifier}(K'_{i,j,k},c_{i,j,k}),\qquad\tilde{r}_{i,j,k}=2\tilde{h}_{i,j,k}-1$. We use the difference between the two signed scores to measure the degree to which the judgement depends on the relevant evidence, $\Delta_{i,j,k}=\left|r_{i,j,k}-\tilde{r}_{i,j,k}\right|$. A large  score change indicates that the original judgement rests on identifiable evidence and is  more reliable, whereas a small score change suggests the verifier reaches a similar judgement even without the key evidence, implying the judgement is more likely driven by linguistic priors or irrelevant information.

We then map the evidence-dependency strength $\Delta_{i,j,k}$ to a continuous reliability weight $w_{i,j,k}=\operatorname{sigmoid}\left(\frac{\Delta_{i,j,k}-\mu}{\tau}\right)\in(0,1)$, where $\mu$ is the median of $\Delta$ computed on a calibration set sampled from the training set, and $\tau$ controls the smoothness of the mapping curve. Finally, the reliability-weighted factual score of each atomic fact is defined as $\tilde{r}_{i,j,k}^{\mathrm{fact}}=w_{i,j,k}r_{i,j,k}$, where $r_{i,j,k}$ determines the direction of the factual supervision signal and $w_{i,j,k}$ determines its strength.

\subsection{Reward Design}
\label{sec:reward-design}

For each rollout $y_i=(y_i^{\mathrm{think}},y_i^{\mathrm{answer}})$, we define three reward components as follows.

\textbf{Format reward} is used to constrain the output structure, verifying whether $y_i$ conforms to the specified format (\texttt{<think>...</think><answer>...</answer>}).
\[
R_i^{\mathrm{format}}
=
\begin{cases}
+1,
&
\text{\small valid format}\\
-1,
&
\text{\small invalid format}\\
\end{cases}
\]

\textbf{Answer reward} is used to evaluate whether the final answer $y_i^{\mathrm{answer}}$ is correct.
\[
R_i^{\mathrm{answer}}
=
\begin{cases}
+1,
&
\text{\small correct answer}\\
-1,
&
\text{\small incorrect answer}\\
\end{cases}
\]

\textbf{Reliability-weighted factual reward} characterizes the degree of support between verifiable facts in $y_i^{\mathrm{think}}$ and the evidence $K$, defined as the average of the reliability-weighted factual scores over all verifiable atomic facts in $y_i^{\mathrm{think}}$.
\[
R_i^{\mathrm{fact}}
=
\begin{cases}
\dfrac{1}{|\mathcal{C}_i|}
\sum_{c_{i,j,k}\in\mathcal{C}_i}
\tilde{r}_{i,j,k}^{\mathrm{fact}},
&
|\mathcal{C}_i|>0,\\
0,
&
|\mathcal{C}_i|=0.
\end{cases}
\]

The final reward for rollout $y_i$ is defined as:
\[
R_i
=
R_i^{\mathrm{format}}
+
R_i^{\mathrm{answer}}
+
R_i^{\mathrm{fact}}.
\]

\subsection{Reliable Fact-Guided Advantage Reshaping}
\label{sec:advantage-reshaping}

FARCA first computes the advantage of rollout $y_i$ following Group Relative Policy Optimization (GRPO)~\cite{shao2024grpo}, and then leverages the alignment between atomic facts and tokens to perform reliability-guided advantage reshaping.

Specifically, for the $G$ rollouts sampled for the same question, we first perform intra-group normalization in the standard GRPO manner to obtain the advantage $A_i=\frac{R_i-\mathrm{mean}(\{R_g\}_{g=1}^{G})}{\mathrm{std}(\{R_g\}_{g=1}^{G})+\epsilon_{\mathrm{std}}}$, which normalizes the reward of $y_i$ relative to the group and indicates whether it should be reinforced or suppressed.
FARCA then incorporates both credit localization and credit reliability into advantage reshaping: credit localization determines which tokens receive the factual signal, while credit reliability determines how strongly the signal is applied to those tokens. This process uses the factual supervision signal to reinforce or correct the relevant tokens.
For each atomic fact, we first use its continuous signed factual score to construct a fact-corrected advantage $A_{i,j,k}^{\mathrm{fact}}=r_{i,j,k}|A_i|$, which is positive when supported by the evidence, negative when conflicting, and zero when neutral. Using $|A_i|$ preserves the relative training strength of the current rollout, while allowing the factual score to determine the update direction for the local tokens.

Furthermore, we use the reliability weight to form a convex combination between the original advantage and the fact-corrected advantage, $\bar{A}_{i,j,k}=(1-w_{i,j,k})A_i+w_{i,j,k}A_{i,j,k}^{\mathrm{fact}}$. When $w_{i,j,k}$ is large, it indicates that the corresponding factual judgement is more strongly evidence-dependent, so the covered tokens rely more on the fact-corrected direction; when $w_{i,j,k}$ is small, the factual judgment is less reliable, and $\bar{A}_{i,j,k}$ remains closer to $A_i$. This soft-flip mechanism avoids hard gradient reversal caused by a single unreliable verification, while still providing positive reinforcement for facts supported by evidence and negative correction for facts that conflict with evidence when the verification outcome is reliable.
The final advantage on token $t$ is:
\[
\hat{A}_{i,t}
=
\begin{cases}
\dfrac{1}{|\mathcal{C}(i,t)|}
\sum_{c_{i,j,k}\in\mathcal{C}(i,t)}
\bar{A}_{i,j,k},
&
\mathcal{C}(i,t)\neq\emptyset,\\
A_i,
&
\mathcal{C}(i,t)=\emptyset.
\end{cases}
\]

where $\mathcal{C}(i,t)=\{c_{i,j,k}:t\in\mathcal{T}(c_{i,j,k})\}$ denotes the set of atomic facts covering token $t$. Tokens covered by multiple facts take their average reliability-aware advantage, and tokens covered by none retain the original.

\subsection{Training Objective}
\label{sec:objective}
Substituting the reliable fact-guided reshaped advantage $\hat{A}_{i,t}$ into the token-level PPO-clip objective, the final optimization objective of FARCA is:
\[
\begin{aligned}
\mathcal{J}_{\mathrm{FARCA}}(\theta)
=&
\ \mathbb{E}_{x,\{y_i\}}
\left[
\frac{1}{G}
\sum_{i=1}^{G}
\frac{1}{|y_i|}
\sum_{t=1}^{|y_i|}
\ell_{i,t}(\theta)
\right]\\
&-
\beta\,
\mathrm{KL}
\big(
\pi_\theta
\Vert
\pi_{\mathrm{ref}}
\big),
\end{aligned}
\]
\[
\ell_{i,t}(\theta)
=
\min
\left(
\rho_{i,t}\hat{A}_{i,t},
\ \mathrm{clip}(\rho_{i,t},1-\epsilon,1+\epsilon)\hat{A}_{i,t}
\right).
\]

where $\rho_{i,t} = \pi_\theta(y_{i,t}\mid x,y_{i,<t}) / \pi_{\theta_{\mathrm{old}}}(y_{i,t}\mid x,y_{i,<t})$ is the token-level importance ratio.

In summary, FARCA mitigates the noisy factual credit assignment problem through fact-token alignment and reliability modeling, injecting factual supervision signals into token-level policy optimization more precisely and robustly.

\section{Experiments}
\label{sec:experiments-zh}

\begin{table*}[t]
\centering
\small
\setlength{\tabcolsep}{1mm}
\renewcommand{\arraystretch}{1.15}
\begin{tabular}{l cccc cccc}
\toprule
\multirow{2}{*}{\textbf{Method}} & \multicolumn{4}{c}{\textbf{Hallucination}} & \multicolumn{4}{c}{\textbf{Math}} \\
\cmidrule(lr){2-5} \cmidrule(lr){6-9}
& \textbf{SimpleQA} & \textbf{TruthfulQA} & \textbf{HalluQA} & \textbf{HaluEval-QA} & \textbf{AIME2026} & \textbf{AIME2025} & \textbf{MATH-500} & \textbf{GSM8K} \\
\midrule

\multicolumn{9}{c}{\textit{\textbf{Qwen2.5-3B-Instruct}}} \\
\midrule
Zero-shot & 1.33 & 30.41 & 16.44 & 23.56 & 0.00 & 0.00 & 32.20 & 65.96 \\
GRPO (Math only)     & 0.79   & 18.12    & 9.56   & 10.22    & \underline{3.33}   & \underline{3.33}   & \underline{54.80}   & \underline{77.18}   \\
GRPO (Full)      & 2.52 & 42.59 & 22.67 & 25.05 & \underline{3.33} & \underline{3.33} & 53.22 & 75.82 \\
KnowRL    & 2.42 & 41.98 & 20.67 & \underline{25.57} & 0.00 & 0.00 & 51.80 & 75.59 \\
FSPO      & 1.60 & 35.49 & 19.33 & 23.74 & 0.00 & \underline{3.33} & 36.20 & 72.10 \\
FaithRL   & \underline{2.63} & \underline{43.32} & \underline{23.11} & 25.36 & \underline{3.33} & 0.00 & 53.80 & 76.12 \\
\textbf{FARCA} & \textbf{3.56} & \textbf{45.41} & \textbf{25.78} & \textbf{26.67} & \textbf{6.67} & \textbf{10.00} & \textbf{63.20} & \textbf{84.46} \\

\midrule

\multicolumn{9}{c}{\textit{\textbf{Llama-3.2-3B-Instruct}}} \\
\midrule
Zero-shot & 1.28 & 22.93 & 7.67 & 21.92 & 0.00 & 0.00 & 27.20 & 60.42 \\
GRPO (Math only)     & 0.67   & 14.20    & 5.33   & 9.53    & \underline{3.33}   & \underline{3.33}   & \underline{42.60}   & \underline{74.22}   \\
GRPO (Full)     & 2.44   & 35.62    & \underline{10.67}   & 23.39    & \underline{3.33}   & \underline{3.33}   & 41.20   & 73.31   \\
KnowRL    & 2.73   & 34.88    & 9.56   & 24.63    & 0.00   & 0.00   & 41.00   & 72.48   \\
FSPO      & 2.09 & 30.05 & 8.45 & 24.14 & 0.00 & 0.00 & 40.40 & 70.36 \\
FaithRL   & \underline{2.81} & \underline{35.74} & 10.00 & \underline{24.89} & 0.00 & 0.00 & 41.80 & 73.92 \\
\textbf{FARCA} & \textbf{3.33} & \textbf{39.05} & \textbf{14.00} & \textbf{25.91} & \textbf{6.67} & \textbf{6.67} & \textbf{47.80} & \textbf{78.24} \\

\bottomrule
\end{tabular}
\caption{Performance comparison on hallucination and math benchmarks. Best results are in bold, and all tied second-best results are underlined.}
\label{tab:main_results}
\end{table*}

\subsection{Experimental Setup}

\paragraph{Datasets and Metrics.}
We use a challenging subset~\cite{song2025r1} of the knowledge-intensive datasets HotpotQA~\cite{yang-etal-2018-hotpotqa} and 2WikiMultiHopQA~\cite{ho-etal-2020-constructing} for FARCA training, where each sample contains a question, an answer, and the corresponding Wikipedia knowledge snippets. In addition, to prevent factuality optimization from sacrificing general reasoning ability, we incorporate the mathematical reasoning dataset SimpleRL~\cite{zeng2025simplerl} for standard RL training.
We use four widely adopted benchmarks, SimpleQA~\cite{wei2024simpleqa}, TruthfulQA~\cite{lin2022truthfulqa}, HalluQA~\cite{cheng2023halluqa}, and HaluEval-QA~\cite{li2023halueval}, for hallucination evaluation. Specifically, we adopt the F1 score for SimpleQA, the truthful ratio for TruthfulQA and HalluQA, and accuracy for HaluEval-QA.
Mathematical reasoning ability is evaluated on AIME2026~\cite{aime2026}, AIME2025~\cite{aime2025}, MATH-500~\cite{hendrycks2021math}, and GSM8K~\cite{cobbe2021gsm8k} under the Pass@1 setting. We uniformly use GPT-4o as the judge model across both hallucination and math benchmarks for truthfulness judgement or answer matching.

\paragraph{Models and Baselines.}
We conduct experiments using Qwen2.5-3B-Instruct~\cite{hui2024qwen25} and Llama-3.2-3B-Instruct~\cite{grattafiori2024llama3}, and consider three categories of baselines. (1) Zero-shot prompting baseline. (2) Outcome-oriented RL methods that use only format and answer rewards. We consider two training-data configurations: GRPO (Math only), trained exclusively on the SimpleRL dataset; and GRPO (Full), trained on the full training dataset. (3) Factual reinforcement learning baselines, including KnowRL~\cite{ren-etal-2026-knowrl}, FSPO~\cite{li2025hallucination}, and FaithRL~\cite{gui2026learning}.

\paragraph{Implementation Details.}
We conduct model training using the verl\footnote{https://github.com/verl-project/verl} framework. We perform full-parameter fine-tuning for 1 epoch on 4 A6000 GPUs, with a learning rate of $5\times10^{-7}$, 6 rollouts per prompt at temperature 1.0, a PPO mini-batch size of 128 (per-GPU batch size 2), a KL coefficient of 0.001, and a maximum prompt length and response length of 2048. We use GPT-4o for atomic fact extraction and token provenance, and the HHEM-2.1-Open model as the fact verifier. For reliability estimation, we remove the evidence sentence most similar to each atomic fact ($K_{\mathrm{rel}}=1$), and map the verification score change to a soft reliability weight using $\tau=0.2$ and center $\mu=0.16$, the median of $\Delta$ over a calibration set sampled from the training data.

\subsection{Main Results}

Table~\ref{tab:main_results} presents the main experimental results.

\paragraph{FARCA consistently improves model factuality while preserving reasoning ability.}
The experimental results show that FARCA improves factuality across different models and hallucination scenarios. On Qwen2.5-3B-Instruct, FARCA achieves the best results on all four hallucination benchmarks; compared to FaithRL, the strongest factuality baseline, FARCA yields an average improvement of 1.75 percentage points, with gains of 2.09 and 2.67 points on TruthfulQA and HalluQA, respectively. On Llama-3.2-3B-Instruct, the average improvement further widens to 2.21 percentage points, with TruthfulQA and HalluQA improving by 3.31 and 4.00 points, respectively. Meanwhile, FARCA also delivers the best performance on the mathematical reasoning benchmarks. This indicates that the gains from FARCA do not stem from incidental fluctuations tied to a single dataset or backbone, but rather from consistently mitigating hallucination across diverse scenarios while simultaneously enhancing the model's reasoning ability.

\begin{table*}[t]
\centering
\small
\setlength{\tabcolsep}{1mm}
\renewcommand{\arraystretch}{1.12}
\begin{tabular}{l cccc cccc}
\toprule
\multirow{2}{*}{\textbf{Variant}} & \multicolumn{4}{c}{\textbf{Hallucination}} & \multicolumn{4}{c}{\textbf{Math}} \\
\cmidrule(lr){2-5} \cmidrule(lr){6-9}
& \textbf{SimpleQA} & \textbf{TruthfulQA} & \textbf{HalluQA} & \textbf{HaluEval-QA} & \textbf{AIME2026} & \textbf{AIME2025} & \textbf{MATH-500} & \textbf{GSM8K} \\
\midrule
\textbf{FARCA} & \textbf{3.56} & \textbf{45.41} & \textbf{25.78} & \textbf{26.67} & \textbf{6.67} & \textbf{10.00} & \textbf{63.20} & \textbf{84.46} \\
w/o token provenance & 3.23 & 44.52 & 23.94 & 25.90 & 6.67 & 6.67 & 61.00 & 82.87 \\
w/o reliability estimation & 3.04 & 43.88 & 24.51 & 26.28 & 3.33 & 6.67 & 61.40 & 82.26 \\
w/o continuous factual score & 3.31 & 44.78 & 25.40 & 26.06 & 6.67 & 10.00 & 62.80 & 83.85 \\
\bottomrule
\end{tabular}
\caption{Component ablation results of FARCA.}
\label{tab:ablation_results}
\end{table*}

\paragraph{The benefit of factual supervision depends on the accuracy and reliability of credit assignment.}
Table~\ref{tab:main_results} shows that although both KnowRL and FSPO explicitly incorporate factual supervision, they underperform standard GRPO (Full) on several evaluation datasets. This suggests that coarse-grained factual rewards may be broadcast across reasoning text containing facts of different factuality, leaving the optimizer unable to distinguish which positions convey correct content and which introduce errors, thereby undermining optimization effectiveness. FaithRL employs step-level supervision with credibility modulation, but it still treats an entire reasoning step as an indivisible update unit. 
The experimental results further demonstrate that the key to leveraging factual supervision lies in converting verification signals into fine-grained and reliable factual credit, thereby enabling more effective and stable guidance for policy optimization.

\subsection{Ablation Studies}
We conduct ablations on Qwen2.5-3B-Instruct.

\paragraph{Component Analysis.}
Component ablations focus on three components. Results are shown in Table~\ref{tab:ablation_results}.
\textbf{Removing token provenance routing (w/o token provenance)} broadcasts factual signals to all tokens within a sentence. This setting drops the average score from 25.36 (FARCA) to 24.40. Results show that restricting factual signals to the tokens corresponding to each atomic fact better handles cases that contain both reliable and hallucinated information simultaneously.
\textbf{Removing counterfactual reliability estimation (w/o reliability estimation)} sets the reliability weight of all factual judgements to 1. This variant also exhibits a clear degradation, suggesting that verifier misjudgements can undermine the benefits of introducing factual supervision and further demonstrating the necessity of reliability estimation for verifier signals.
\textbf{Removing the continuous signed factual score (w/o continuous factual score)} discretizes the raw NLI verifier score into $\{-1,0,+1\}$ using 0.5 as the threshold. This setting causes a modest performance drop, confirming that the continuous score provides a smoother, more fine-grained optimization signal than hard labels for token-level credit assignment.

\begin{figure}[t]
    \centering
    \includegraphics[width=0.95\columnwidth]{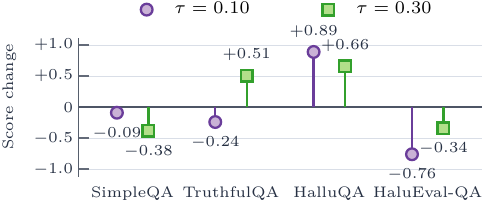}
    \caption{
    Hyperparameter sensitivity of FARCA to the reliability temperature $\tau$ (score change from $\tau=0.20$).
    }
    \label{fig:tau_sensitivity}
\end{figure}

\paragraph{Hyperparameter Sensitivity.}
We further examine the effect of the reliability temperature $\tau$ on FARCA's factuality. Figure~\ref{fig:tau_sensitivity} shows the results on the four hallucination benchmarks for $\tau\in\{0.10,0.20,0.30\}$. The corresponding average scores are 25.31, 25.36, and 25.47, with a maximum difference of only 0.16 percentage points, indicating that FARCA is generally robust to the temperature setting within this range. We adopt $\tau=0.20$ as the default value, which preserves  discrimination while avoiding an overly sharp mapping.

\subsection{Further Analysis}

\begin{figure}[t]
    \centering
    \includegraphics[width=0.95\columnwidth]{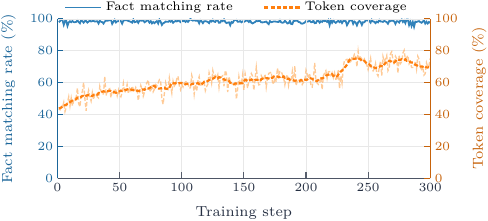}
    \caption{
    Fact matching rate and the percentage of tokens covered by matched fact spans during training.
    }
    \label{fig:fact_token_alignment}
\end{figure}

\paragraph{Fact--token alignment provides stable and broad supervision.}
Figure~\ref{fig:fact_token_alignment} examines whether atomic facts can be reliably routed back to their source token spans. The blue curve (left axis) shows the matching rate: throughout training, about 98.3\% of extracted facts are successfully aligned, and the rate remains consistently high, indicating that FARCA reliably tracks token provenance in practice. The orange curve (right axis) shows the proportion of reasoning tokens covered by matched facts. This proportion rises steadily early in training before fluctuating at a higher level, reflecting that an increasing share of generated content becomes associated with verifiable facts and receives factual supervision. These results show that FARCA achieves both reliable localization and sufficiently broad supervision coverage.

\begin{figure}[t]
    \centering
    \includegraphics[width=0.97\columnwidth]{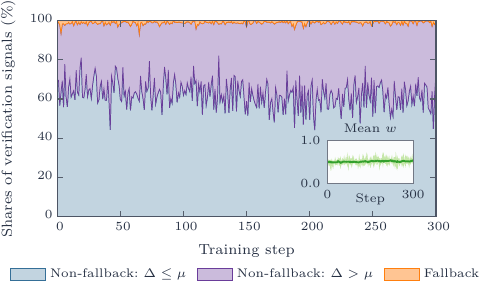}
    \caption{
    Proportions of verification signals grouped by $\Delta$ relative to $\mu$ and by fallback status over FARCA training, with mean continuous weight $w$.
    }
    \label{fig:reliability_separation}
\end{figure}

\paragraph{Counterfactual evidence attribution yields stable, non-degenerate reliability signals.}
Figure~\ref{fig:reliability_separation} examines whether counterfactual evidence attribution produces discriminative reliability signals during training. FARCA estimates the signal's reliability from the verifier score change $\Delta$ after removing its most relevant evidence sentence. The stacked areas show the proportions of facts with $\Delta>\mu$, $\Delta\leq\mu$, and fallback, while the line shows the mean continuous reliability weight $w$. We observe that 36.3\% of facts have $\Delta>\mu$, and that only 1.1\% fallback, with an average weight of 0.512, neither collapsing to 0 nor 1. This shows that counterfactual evidence attribution assigns stable, non-degenerate weights based on each fact's evidence dependence. It effectively down-weights verifier misjudgements that are insensitive to key evidence, reducing their noise during optimization.

\begin{figure}[t]
    \centering
    \includegraphics[width=0.97\columnwidth]{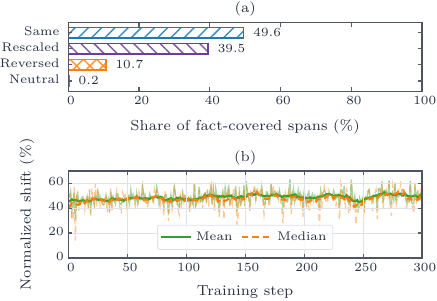}
    \caption{Effect of reliability-aware soft interpolation on the advantages of fact-covered tokens. (a) Directional outcome of the interpolated advantage relative to the original advantage. (b) Per-step mean and median trajectories of the normalized shift over training.}
    \label{fig:advantage_reshaping}
\end{figure}

\paragraph{Reliability-aware soft interpolation substantially reshapes fact-span advantages.}
Figure~\ref{fig:advantage_reshaping} examines whether reliability-aware soft interpolation actually injects fact-level feedback into the span advantages assigned to fact-covered tokens before overlapping-fact aggregation.
Plot (a) summarizes the directional outcome across fact-covered spans: 49.6\% retain direction, 39.5\% retain direction but with rescaled magnitude, and 10.7\% reverse (0.2\% neutral), indicating that FARCA continuously mediates between preserving global credit and applying local correction.
Plot (b) shows that the span-count-weighted average of the per-step mean normalized shift is 48.8\%, while the per-step median tracks the mean closely throughout training.
These results provide evidence that soft interpolation converts coarse sequence credit into localized fact-span signals that incorporate factual feedback and are subsequently aggregated into the final token advantage.

\begin{figure}[t]
    \centering
    \includegraphics[width=0.97\columnwidth]{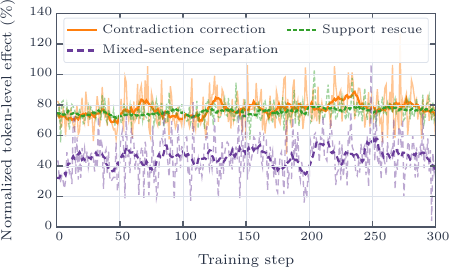}
    \caption{
    Normalized token-level correction of local--global credit conflicts during FARCA training.
    }
    \label{fig:local_global_correction}
\end{figure}

\paragraph{Token Provenance Enables Local--Global Credit Correction.}
Figure~\ref{fig:local_global_correction} examines whether token provenance can adjust global advantages that conflict with local factual feedback. Contradiction correction and support rescue average 77.1\% and 75.6\% in strength, reliably flipping token-level advantages that conflict with the rollout's overall sign. Moreover, mixed-sentence separation reaches 45.9\%, showing that even correct and incorrect facts within the same sentence receive clearly distinct final advantages.  Overall, FARCA continuously modulates global advantage by the direction and reliability of local facts, curbing erroneous credit assignment while preserving effective global learning signals for more precise and robust optimization.

\section{Conclusion}

We proposed FARCA to address noisy factual credit assignment in RL with factual supervision. It resolves credit localization ambiguity via fact–token alignment, which matches verification granularity to policy optimization, and credit reliability ambiguity via counterfactual evidence attribution, which estimates signal reliability to guide reward computation and advantage reshaping. This transforms noisy, unaligned verification signals into fine-grained, reliability-aware credit, allowing factual supervision to be incorporated into policy optimization more accurately and robustly. Experiments confirm FARCA's effectiveness and stability in improving model factuality.

\bibliography{aaai2027}

@misc{jaech2024openai,
  title={Openai o1 system card},
  author={Jaech, Aaron and Kalai, Adam and Lerer, Adam and Richardson, Adam and El-Kishky, Ahmed and Low, Aiden and Helyar, Alec and Madry, Aleksander and Beutel, Alex and Carney, Alex and others},
  eprint={2412.16720},
  archivePrefix={arXiv},
  year={2024}
}

@misc{guo2025deepseek,
  title={Deepseek-r1: Incentivizing reasoning capability in llms via reinforcement learning},
  author={Guo, Daya and Yang, Dejian and Zhang, Haowei and Song, Junxiao and Wang, Peiyi and Zhu, Qihao and Xu, Runxin and Zhang, Ruoyu and Ma, Shirong and Bi, Xiao and others},
  eprint={2501.12948},
  archivePrefix={arXiv},
  year={2025}
}

@article{huang2025survey,
  title={A survey on hallucination in large language models: Principles, taxonomy, challenges, and open questions},
  author={Huang, Lei and Yu, Weijiang and Ma, Weitao and Zhong, Weihong and Feng, Zhangyin and Wang, Haotian and Chen, Qianglong and Peng, Weihua and Feng, Xiaocheng and Qin, Bing and others},
  journal={ACM Transactions on Information Systems},
  year={2025}
}

@article{zhang2025siren,
  title={Siren’s Song in the AI Ocean: A Survey on Hallucination in Large Language Models},
  author={Zhang, Yue and Li, Yafu and Cui, Leyang and Cai, Deng and Liu, Lemao and Fu, Tingchen and Huang, Xinting and Zhao, Enbo and Zhang, Yu and Chen, Yulong and others},
  journal={Computational Linguistics},
  year={2025}
}

@misc{gui2026learning,
  title={Learning to Reason Faithfully through Step-Level Faithfulness Maximization},
  author={Gui, Runquan and Li, Yafu and Qu, Xiaoye and Liu, Ziyan and Cheng, Yeqiu and Cheng, Yu},
  eprint={2602.03507},
  archivePrefix={arXiv},
  year={2026}
}

@inproceedings{ren-etal-2026-knowrl,
    title = "{K}now{RL}: Exploring Knowledgeable Reinforcement Learning for Factuality",
    author = "Ren, Baochang  and
      Qiao, Shuofei  and
      Zhang, Ningyu  and
      Zheng, Da  and
      Chen, Huajun",
    booktitle = "Proceedings of the 64th Annual Meeting of the ACL (Volume 1: Long Papers)",
    year = "2026"
}

@inproceedings{li2025hallucination,
  title={Reasoning models hallucinate more: Factuality-aware reinforcement learning for large reasoning models},
  author={Li, Junyi and Ng, Hwee Tou},
  booktitle={Advances in Neural Information Processing Systems},
  volume={38},
  pages={151064--151085},
  year={2025}
}

@inproceedings{chen-etal-2025-reference,
    title = "On Reference (In-)Determinacy in Natural Language Inference",
    author = "Chen, Sihao  and
      Malaviya, Chaitanya  and
      Fabrikant, Alex  and
      Taitelbaum, Hagai  and
      Schuster, Tal  and
      Buthpitiya, Senaka  and
      Roth, Dan",
    booktitle = "Findings of the Association for Computational Linguistics: NAACL 2025",
    year = "2025"
}

@misc{cai2025reinforcement,
  title={Reinforcement learning with verifiable yet noisy rewards under imperfect verifiers},
  author={Cai, Xin-Qiang and Wang, Wei and Liu, Feng and Liu, Tongliang and Niu, Gang and Sugiyama, Masashi},
  eprint={2510.00915},
  archivePrefix={arXiv},
  year={2025}
}

@misc{ding2025fapo,
  title={FAPO: flawed-aware policy optimization for efficient and reliable reasoning},
  author={Ding, Yuyang and Zhang, Chi and Li, Juntao and Lin, Haibin and Zhang, Min},
  eprint={2510.22543},
  archivePrefix={arXiv},
  year={2025}
}

@misc {hhem-2.1-open,
    author       = {Miaoran Li and Rogger Luo and Ofer Mendelevitch},
    title        = {{HHEM-2.1-Open}},
    year         = 2024,
    url          = { https://huggingface.co/vectara/hallucination_evaluation_model },
    doi          = { 10.57967/hf/3240 },
    publisher    = { Hugging Face }
}

@inproceedings{zha-etal-2023-alignscore,
    title = "{A}lign{S}core: Evaluating Factual Consistency with A Unified Alignment Function",
    author = "Zha, Yuheng  and
      Yang, Yichi  and
      Li, Ruichen  and
      Hu, Zhiting",
    booktitle = "Proceedings of the 61st Annual Meeting of the ACL (Volume 1: Long Papers)",
    year = "2023"
}

@inproceedings{tang-etal-2024-minicheck,
    title = "{M}ini{C}heck: Efficient Fact-Checking of {LLM}s on Grounding Documents",
    author = "Tang, Liyan  and
      Laban, Philippe  and
      Durrett, Greg",
    booktitle = "Proceedings of the 2024 Conference on EMNLP",
    year = "2024"
}

@inproceedings{ho-etal-2020-constructing,
    title = "Constructing A Multi-hop {QA} Dataset for Comprehensive Evaluation of Reasoning Steps",
    author = "Ho, Xanh  and
      Duong Nguyen, Anh-Khoa  and
      Sugawara, Saku  and
      Aizawa, Akiko",
    booktitle = "Proceedings of the 28th International Conference on Computational Linguistics",
    year = "2020"
}

@inproceedings{yang-etal-2018-hotpotqa,
    title = "{H}otpot{QA}: A Dataset for Diverse, Explainable Multi-hop Question Answering",
    author = "Yang, Zhilin  and
      Qi, Peng  and
      Zhang, Saizheng  and
      Bengio, Yoshua  and
      Cohen, William  and
      Salakhutdinov, Ruslan  and
      Manning, Christopher D.",
    booktitle = "Proceedings of the 2018 Conference on EMNLP",
    year = "2018"
}

@misc{song2025r1,
  title={R1-searcher: Incentivizing the search capability in llms via reinforcement learning},
  author={Song, Huatong and Jiang, Jinhao and Min, Yingqian and Chen, Jie and Chen, Zhipeng and Zhao, Wayne Xin and Fang, Lei and Wen, Ji-Rong},
  eprint={2503.05592},
  archivePrefix={arXiv},
  year={2025}
}

@misc{zeng2025simplerl,
  title={7b model and 8k examples: Emerging reasoning with reinforcement learning is both effective and efficient},
  author={Zeng, Weihao and Huang, Yuzhen and Liu, Wei and He, Keqing and Liu, Qian and Ma, Zejun and He, Junxian},
  year={2025}
}

@misc{wei2024simpleqa,
  title={Measuring short-form factuality in large language models},
  author={Wei, Jason and Karina, Nguyen and Chung, Hyung Won and Jiao, Yunxin Joy and Papay, Spencer and Glaese, Amelia and Schulman, John and Fedus, William},
  eprint={2411.04368},
  archivePrefix={arXiv},
  year={2024}
}

@misc{cheng2023halluqa,
  title={Evaluating hallucinations in chinese large language models},
  author={Cheng, Qinyuan and Sun, Tianxiang and Zhang, Wenwei and Wang, Siyin and Liu, Xiangyang and Zhang, Mozhi and He, Junliang and Huang, Mianqiu and Yin, Zhangyue and Chen, Kai and others},
  eprint={2310.03368},
  archivePrefix={arXiv},
  year={2023}
}

@inproceedings{li2023halueval,
  title={Halueval: A large-scale hallucination evaluation benchmark for large language models},
  author={Li, Junyi and Cheng, Xiaoxue and Zhao, Xin and Nie, Jian-Yun and Wen, Ji-Rong},
  booktitle={The 2023 Conference on EMNLP},
  year={2023}
}

@inproceedings{lin2022truthfulqa,
  title={Truthfulqa: Measuring how models mimic human falsehoods},
  author={Lin, Stephanie and Hilton, Jacob and Evans, Owain},
  booktitle={Proceedings of the 60th annual meeting of the ACL (volume 1: long papers)},
  year={2022}
}

@misc{cobbe2021gsm8k,
  title={Training verifiers to solve math word problems},
  author={Cobbe, Karl and Kosaraju, Vineet and Bavarian, Mohammad and Chen, Mark and Jun, Heewoo and Kaiser, Lukasz and Plappert, Matthias and Tworek, Jerry and Hilton, Jacob and Nakano, Reiichiro and others},
  eprint={2110.14168},
  archivePrefix={arXiv},
  year={2021}
}

@inproceedings{hendrycks2021math,
  title={Measuring mathematical problem solving with the math dataset},
  author={Hendrycks, Dan and Burns, Collin and Kadavath, Saurav and Arora, Akul and Basart, Steven and Tang, Eric and Song, Dawn and Steinhardt, Jacob},
  booktitle = "Proceedings of the Thirty-fifth Annual Conference on NeurIPS, 2021.",
  year={2021}
}

@misc{aime2025,
  author       = {{MAA}},
  title        = {American Invitational Mathematics Examination - {AIME} 2025},
  year         = {2025}
}

@misc{aime2026,
  author       = {{MAA}},
  title        = {American Invitational Mathematics Examination - {AIME} 2026},
  year         = {2026}
}

@misc{hurst2024gpt4o,
  title={Gpt-4o system card},
  author={Hurst, Aaron and Lerer, Adam and Goucher, Adam P and Perelman, Adam and Ramesh, Aditya and Clark, Aidan and Ostrow, AJ and Welihinda, Akila and Hayes, Alan and Radford, Alec and others},
  eprint={2410.21276},
  archivePrefix={arXiv},
  year={2024}
}

@misc{hui2024qwen25,
  title={Qwen2. 5-coder technical report},
  author={Hui, Binyuan and Yang, Jian and Cui, Zeyu and Yang, Jiaxi and Liu, Dayiheng and Zhang, Lei and Liu, Tianyu and Zhang, Jiajun and Yu, Bowen and Lu, Keming and others},
  eprint={2409.12186},
  archivePrefix={arXiv},
  year={2024}
}

@misc{grattafiori2024llama3,
  title={The llama 3 herd of models},
  author={Grattafiori, Aaron and Dubey, Abhimanyu and Jauhri, Abhinav and Pandey, Abhinav and Kadian, Abhishek and Al-Dahle, Ahmad and Letman, Aiesha and Mathur, Akhil and Schelten, Alan and Vaughan, Alex and others},
  eprint={2407.21783},
  archivePrefix={arXiv},
  year={2024}
}

@misc{shao2024grpo,
  title={Deepseekmath: Pushing the limits of mathematical reasoning in open language models},
  author={Shao, Zhihong and Wang, Peiyi and Zhu, Qihao and Xu, Runxin and Song, Junxiao and Bi, Xiao and Zhang, Haowei and Zhang, Mingchuan and Li, YK and Wu, Yang and others},
  eprint={2402.03300},
  archivePrefix={arXiv},
  year={2024}
}

@misc{chen2025learning,
  title={Learning to reason for factuality},
  author={Chen, Xilun and Kulikov, Ilia and Berges, Vincent-Pierre and O{\u{g}}uz, Barlas and Shao, Rulin and Ghosh, Gargi and Weston, Jason and Yih, Wen-tau},
  eprint={2508.05618},
  archivePrefix={arXiv},
  year={2025}
}

@misc{chen2025reasoning,
  title={Reasoning models don't always say what they think},
  author={Chen, Yanda and Benton, Joe and Radhakrishnan, Ansh and Uesato, Jonathan and Denison, Carson and Schulman, John and Somani, Arushi and Hase, Peter and Wagner, Misha and Roger, Fabien and others},
  eprint={2505.05410},
  archivePrefix={arXiv},
  year={2025}
}

@inproceedings{paul2024making,
  title={Making reasoning matter: Measuring and improving faithfulness of chain-of-thought reasoning},
  author={Paul, Debjit and West, Robert and Bosselut, Antoine and Faltings, Boi},
  booktitle={Findings of the Association for Computational Linguistics: EMNLP 2024},
  year={2024}
}

@inproceedings{wang2026joint,
  title={Joint evaluation of answer and reasoning consistency for hallucination detection in large reasoning models},
  author={Wang, Changyue and Su, Weihang and Ai, Qingyao and Liu, Yiqun},
  booktitle={Proceedings of the AAAI Conference on Artificial Intelligence},
  year={2026}
}

\clearpage

\appendix

\section{Implementation Details}

\subsection{Datasets}

\paragraph{Training Data.}
Our training data combines knowledge-intensive question-answering examples with mathematical reasoning problems. For factuality-oriented training, we adopt the challenging subset~\cite{song2025r1} constructed from HotpotQA~\cite{yang-etal-2018-hotpotqa} and 2WikiMultiHopQA~\cite{ho-etal-2020-constructing}. This collection contains 4,761 samples from HotpotQA and 3,737 samples from 2WikiMultiHopQA, spanning varying levels of difficulty. Each sample consists of a question, its ground-truth answer, and the corresponding Wikipedia knowledge snippets, thereby providing explicit evidence for factuality-aware optimization. We use a subset of 2,000 examples randomly sampled from this collection~\cite{li2025hallucination} for FARCA training. To preserve the model's general reasoning ability during factuality optimization, we additionally incorporate SimpleRL~\cite{zeng2025simplerl}, which contains 8,523 mathematical reasoning problems, and use these examples for standard reinforcement learning with GRPO.

\paragraph{Evaluation Data.}
We evaluate hallucinations using four widely adopted datasets: SimpleQA~\cite{wei2024simpleqa}, TruthfulQA~\cite{lin2022truthfulqa}, HalluQA~\cite{cheng2023halluqa}, and HaluEval-QA~\cite{li2023halueval}. SimpleQA is one of the most challenging short-form factual question-answering benchmarks, assessing a model’s precise recall of long-tail facts using the F1 score. TruthfulQA elicits untruthful responses through common misconceptions and false beliefs, and uses the truthful ratio to measure a model’s ability to resist producing incorrect answers. HalluQA is designed for Chinese hallucination evaluation and covers adversarial questions across multiple domains, including topics related to Chinese culture; it likewise reports the truthful ratio. Given relevant knowledge, HaluEval-QA contrasts correct answers with hallucinated ones and uses accuracy to evaluate a model’s ability to identify fine-grained factual inconsistencies. For answer matching and truthfulness assessment in hallucination evaluation, we use GPT-4o as the judge model to determine whether an output is correct or truthful.
Mathematical reasoning ability is evaluated on AIME2026~\cite{aime2026}, AIME2025~\cite{aime2025}, MATH-500~\cite{hendrycks2021math}, and GSM8K~\cite{cobbe2021gsm8k}. We adopt the standard Pass@1 setting and use GPT-4o to determine whether model outputs exactly match or are symbolically equivalent to the reference answers.

\subsection{Component Ablation Settings}

All component ablations follow the same experimental configuration as full FARCA and modify only the component under investigation.

\textbf{w/o token provenance.}
This variant removes the token provenance $\mathcal{T}(c_{i,j,k})$ used to route each factual signal back to its source tokens. The atomic facts, continuous signed factual scores, reliability weights, and fact-level reliability-aware advantages $\bar{A}_{i,j,k}$ are computed in the same way as in full FARCA. However, instead of assigning $\bar{A}_{i,j,k}$ only to the tokens covered by $c_{i,j,k}$, we average the advantages of all atomic facts extracted from the same sentence:
\[
A_{i,j}^{\mathrm{sent}}
=
\frac{1}{N_{i,j}}
\sum_{k=1}^{N_{i,j}}
\bar{A}_{i,j,k}.
\]
Let $\mathcal{T}(s_{i,j})$ denote all token positions belonging to sentence $s_{i,j}$. The resulting sentence-level advantage is broadcast to every token in that sentence:
\[
\hat{A}_{i,t}
=
\begin{cases}
A_{i,j}^{\mathrm{sent}},
&
t\in\mathcal{T}(s_{i,j}),\ N_{i,j}>0,\\
A_i,
&
\text{otherwise}.
\end{cases}
\]
The reliability-weighted factual reward $R_i^{\mathrm{fact}}$ remains unchanged. Thus, this variant removes only the localization of factual credit while retaining the same fact-level supervision signals.

\textbf{w/o reliability estimation.}
This variant bypasses counterfactual evidence attribution and treats every verification outcome as fully reliable by setting
\[
w_{i,j,k}=1
\]
for all atomic facts. Accordingly, the reliability-weighted factual score reduces to
\[
\tilde{r}_{i,j,k}^{\mathrm{fact}}
=
r_{i,j,k},
\]
and the reliability-aware advantage reduces to the fact-corrected advantage:
\[
\bar{A}_{i,j,k}
=
A_{i,j,k}^{\mathrm{fact}}
=
r_{i,j,k}|A_i|.
\]
Atomic fact extraction and token provenance are retained, so each fact-corrected advantage is still assigned only to the tokens in $\mathcal{T}(c_{i,j,k})$, following the same token-level aggregation rule as full FARCA. This variant therefore isolates the contribution of estimating the reliability of verifier judgments.

\textbf{w/o continuous factual score.}
This variant preserves token provenance and counterfactual reliability estimation but replaces the continuous signed factual score $r_{i,j,k}=2h_{i,j,k}-1$ with a discrete score obtained by thresholding the verifier output $h_{i,j,k}$ at 0.5:
\[
r_{i,j,k}^{\mathrm{disc}}
=
\begin{cases}
+1, & h_{i,j,k}>0.5,\\
0,  & h_{i,j,k}=0.5,\\
-1, & h_{i,j,k}<0.5.
\end{cases}
\]
To isolate the effect of score discretization, the reliability weight $w_{i,j,k}$ is estimated exactly as in full FARCA using the continuous original and counterfactual verification scores. The discrete score replaces $r_{i,j,k}$ only when constructing the factual reward and fact-corrected advantage:
\[
\tilde{r}_{i,j,k}^{\mathrm{fact}}
=
w_{i,j,k}r_{i,j,k}^{\mathrm{disc}},
\qquad
A_{i,j,k}^{\mathrm{fact}}
=
r_{i,j,k}^{\mathrm{disc}}|A_i|.
\]
The reliability-aware advantage is subsequently computed as
\[
\bar{A}_{i,j,k}
=
(1-w_{i,j,k})A_i
+
w_{i,j,k}A_{i,j,k}^{\mathrm{fact}},
\]
with the same token provenance routing and aggregation rule as full FARCA. Therefore, this variant removes only the continuous magnitude of the factual supervision signal while preserving its direction and reliability calibration.

\subsection{Prompt Templates}

Table~\ref{tab:atomic_fact_extraction_prompt} presents the prompt used for atomic fact extraction and token provenance localization. For each verifiable sentence, GPT-4o returns a set of self-contained atomic facts, each paired with a minimal textual \texttt{source\_span} that exactly matches the corresponding substring of the original sentence; sentences without verifiable factual content yield an empty set. We subsequently align each source span with the tokenized rollout to obtain the token index set $\mathcal{T}(c_{i,j,k})$, ensuring that each factual signal is routed only to the tokens responsible for expressing the corresponding fact.

\begin{table*}[t]
\centering
{%
\small
\setlength{\tabcolsep}{3pt}
\begin{tabular}{p{\dimexpr\textwidth-2\tabcolsep\relax}}
\toprule

\textbf{Instruction:}\par
You are an atomic fact extractor. Given a SENTENCE, extract all atomic
facts from it. An atomic fact is the smallest, self-contained, and
independently verifiable factual statement.\par
1. Each atomic fact must be a complete and self-contained statement
that can be understood independently.\par
2. Remove subjective opinions, speculation, reasoning steps, and
discourse connectors; only keep factual content.\par
3. If the SENTENCE does not contain any verifiable facts
(e.g., pure reasoning, mathematical derivations, or subjective
judgments), return an empty list.\par
4. For each atomic fact, provide its corresponding
\texttt{source\_span}, which must be an exact substring of the
SENTENCE.\par
5. Atomic facts should be ordered according to their appearance in
the SENTENCE.\par
6. The \texttt{source\_span} should include only the minimal,
distinctive part corresponding to that fact. Avoid repeating shared
subjects or prefixes across multiple facts.\par
7. Your task is to do this for the SENTENCE under ``Your Task.''
Some examples have been provided for you to learn how to do this task.

\par\medskip
EXAMPLE \#1:\par
SENTENCE:
Beijing is the capital of the United States, Tiananmen is one of the
most famous landmarks of Beijing.

\par\smallskip
ATOMIC FACTS:\par
\texttt{\textasciigrave\textasciigrave\textasciigrave{}json}\par
\texttt{\{"atomic\_facts": [\{"fact": "Beijing is the capital of the
United States", "source\_span": "Beijing is the capital of the United
States"\}, \{"fact": "Tiananmen is one of the most famous landmarks
of Beijing", "source\_span": "Tiananmen is one of the most famous
landmarks of Beijing"\}]\}}\par
\texttt{\textasciigrave\textasciigrave\textasciigrave}

\par\medskip
EXAMPLE \#2:\par
SENTENCE:
Therefore, we can conclude that this reasoning is correct.

\par\smallskip
ATOMIC FACTS:\par
\texttt{\textasciigrave\textasciigrave\textasciigrave{}json}\par
\texttt{\{"atomic\_facts": []\}}\par
\texttt{\textasciigrave\textasciigrave\textasciigrave}

\par\medskip
EXAMPLE \#3:\par
SENTENCE:
Tiananmen Square in Beijing is the largest city square in the world
and attracts millions of visitors each year.

\par\smallskip
ATOMIC FACTS:\par
\texttt{\textasciigrave\textasciigrave\textasciigrave{}json}\par
\texttt{\{"atomic\_facts": [\{"fact": "Tiananmen Square in Beijing is
the largest city square in the world", "source\_span": "Tiananmen
Square in Beijing is the largest city square in the world"\},
\{"fact": "Tiananmen Square in Beijing attracts millions of visitors
each year", "source\_span": "attracts millions of visitors each
year"\}]\}}\par
\texttt{\textasciigrave\textasciigrave\textasciigrave}

\par\medskip
EXAMPLE \#4:\par
SENTENCE:
Beijing is the capital of the United States, its landmark building is
Tiananmen.

\par\smallskip
ATOMIC FACTS:\par
\texttt{\textasciigrave\textasciigrave\textasciigrave{}json}\par
\texttt{\{"atomic\_facts": [\{"fact": "Beijing is the capital of the
United States", "source\_span": "Beijing is the capital of the United
States"\}, \{"fact": "Beijing's landmark building is Tiananmen",
"source\_span": "its landmark building is Tiananmen"\}]\}}\par
\texttt{\textasciigrave\textasciigrave\textasciigrave}

\par\medskip
EXAMPLE \#5:\par
SENTENCE:
Marie Curie was a famous chemist, physicist, and writer.

\par\smallskip
ATOMIC FACTS:\par
\texttt{\textasciigrave\textasciigrave\textasciigrave{}json}\par
\texttt{\{"atomic\_facts": [\{"fact": "Marie Curie was a famous
chemist", "source\_span": "Marie Curie was a famous chemist"\},
\{"fact": "Marie Curie was a famous physicist", "source\_span":
"physicist"\}, \{"fact": "Marie Curie was a famous writer",
"source\_span": "writer"\}]\}}\par
\texttt{\textasciigrave\textasciigrave\textasciigrave}

\par\medskip
Your Task:\par
SENTENCE:
\texttt{[SENTENCE]}

\par\smallskip
ATOMIC FACTS:

\\
\bottomrule
\end{tabular}
}%
\caption{Prompt used for atomic fact extraction and source-span
localization. The placeholder \texttt{[SENTENCE]} is replaced with
each verifiable sentence in the reasoning trajectory.}
\label{tab:atomic_fact_extraction_prompt}
\end{table*}


\end{document}